\documentclass[conference]{IEEEtran}
\IEEEoverridecommandlockouts
\usepackage{cite}
\usepackage{amsmath,amssymb,amsfonts}
\usepackage{algorithmic}
\usepackage{graphicx}
\usepackage{textcomp}
\usepackage{xcolor}
\def\BibTeX{{\rm B\kern-.05em{\sc i\kern-.025em b}\kern-.08em
    T\kern-.1667em\lower.7ex\hbox{E}\kern-.125emX}}

\begin{document}

\title{Black-box Adversarial Attacks in \\ Autonomous Vehicle Technology
}

\author{
K. Naveen Kumar\textsuperscript{1},
C. Vishnu\textsuperscript{1},
Reshmi Mitra\textsuperscript{2}, 
C. Krishna Mohan\textsuperscript{1}\\
\textsuperscript{1} Indian Institute of Technology Hyderabad, India \\
\textsuperscript{2} Southeast Missouri State University, Cape Girardeau, USA \\
\small{\{cs19m20p000001, cs16m18p000001\}@iith.ac.in, rmitra@semo.edu, ckm@cse.iith.ac.in}
}

\IEEEoverridecommandlockouts
\IEEEpubid{\makebox[\columnwidth]{978-1-7281-8243-8/20/\$31.00~\copyright2020 IEEE \hfill} \hspace{\columnsep}\makebox[\columnwidth]{ }}

\maketitle

\IEEEpubidadjcol

\begin{abstract}
Despite the high quality performance of the deep neural network in real-world applications, they are susceptible to minor perturbations of adversarial attacks. This is mostly undetectable to human vision. The impact of such attacks has become extremely detrimental in autonomous vehicles with real-time “safety” concerns. The black-box adversarial attacks cause drastic misclassification in critical scene elements such as road signs and traffic lights leading the autonomous vehicle to crash into other vehicles or pedestrians. In this paper, we propose a novel query-based attack method called Modified Simple black-box attack (M-SimBA) to overcome the use of a white-box source in transfer based attack method. Also, the issue of late convergence in a Simple black-box attack (SimBA) is addressed by minimizing the loss of the most confused class which is the incorrect class predicted by the model with the highest probability, instead of trying to maximize the loss of the correct class. We evaluate the performance of the proposed approach to the German Traffic Sign Recognition Benchmark (GTSRB) dataset. We show that the proposed model outperforms the existing models like Transfer-based projected gradient descent (T-PGD), SimBA in terms of convergence time, flattening the distribution of confused class probability, and producing adversarial samples with least confidence on the true class.

\end{abstract}

\begin{IEEEkeywords}
adversarial attacks, black-box attacks, deep learning methods, autonomous vehicles.
\end{IEEEkeywords}

\section{Introduction}
\noindent Cybersecurity threats on Autonomous vehicles (AV) can cause serious safety and security issues as per the ``Safety First'' industry consortium paper~\cite{wood2019safety} published by twelve industry leaders such as Audi, BMW, Volkswagen, among others. AV is made possible due to the control functions of connected vehicles, onboard diagnostics for maintenance, and cloud backend system. These capabilities also make it a rich and vulnerable attack surface for the adversary. Cyber-attacks on such systems can have dangerous effects leading to malicious actors gaining arbitrary control of the vehicle with such multiple entities managed simultaneously on the road. These malicious actions can eventually cause life-threatening harm to pedestrians and prevent widespread adoption of AV.

\begin{figure}[htb]
    \centering
    \includegraphics[scale = 0.4]{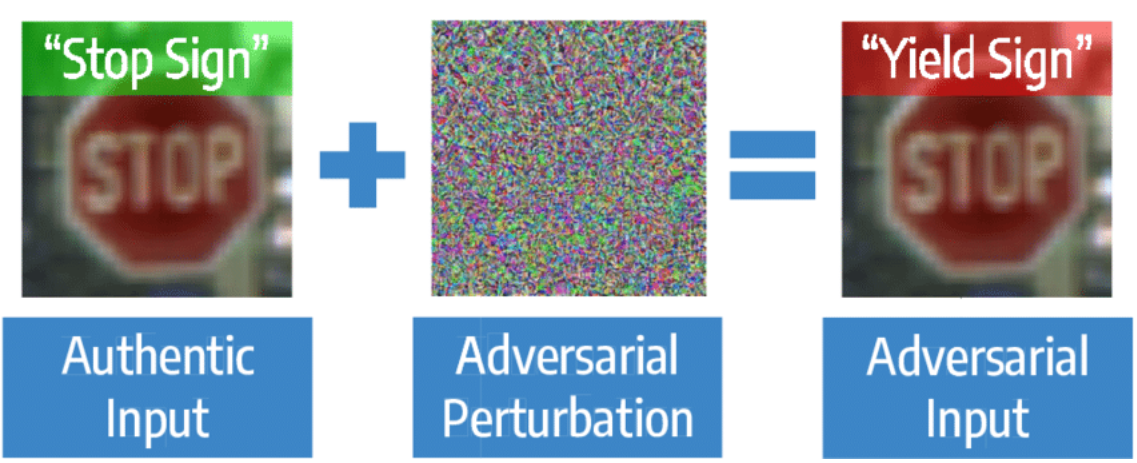}
    \caption{Example of adversarial attack: minor perturbations introduced to the training data cause misclassification of a critical traffic sign i.e. Yield instead of  Stop sign. This incorrect prediction can be hardly perceptible to the human eye and thus have dangerous repercussions for autonomous vehicles.}
    \label{fig:adversarial_1}
\end{figure}

Cyber attacks often cause data corruption and intentional tampering by an unexpected source, which could be crucial elements in the training data for deep neural networks \cite{deng2020analysis}. Although these models are popular for their accuracy and performance for computer vision tasks (such as classification, detection, and segmentation), they are known to be extremely vulnerable to adversarial attacks \cite{X}. In this type of attack, the adversary induces minor but systematic perturbations in key model layers such as filters and input datasets as shown in Fig. \ref{fig:adversarial_1}. Even though this minor layer of noise is barely perceptible to human vision, it may cause drastic misclassification in critical scene elements such as road signs and traffic lights. This may eventually lead to AV crashing into other vehicles or pedestrians. Stickers or paintings on the traffic signboards are the most common physical adversarial attacks, which can impact the functionality of the vehicular system.

Adversarial attacks are primarily of two types: (1) \textit{White-box} where adversary customizes perturbations to the known deep neural network such as architecture, training data, parameter settings, and (2) \textit{Black-box} where adversary has minimum to nil knowledge about the network. Although white-box attacks have been under study, they may not be realistic for AV technology, because of  the many dynamic elements primarily related to sensor data. Our state-of-art study has shown that there is very limited research on black-box adversarial attacks in the domain of AV.

Seminal research articles \cite{X, Y} to report adversarial attack problems for images in neural networks observed that an imperceptible non-random noise to a test image can lead to serious misprediction problems, thereby questioning the model robustness. These white box examples were generated using box-constrained  Limited-memory Broyden–Fletcher–Goldfarb–Shanno (L-BFGS) algorithm. It has remarkable transferability property and is illustrated across tasks with different architectures \cite{P, Q}. 
The decision outputs resulted from machine learning models of the sub-tasks in the computer vision domain, such as classification, detection, and segmentation, become sensitive to the adversarial perturbations in the input. This is discussed in various prior works \cite{A, B, C, D}. 

Gradient estimation techniques such as Finite Differences (FD) and Natural Evolutionary Strategies (NES) are used in a black-box setting, because they are not directly accessible to the adversary. The other significant technique uses surrogates ~\cite{papernot2016transferability, 
moosavi2017pascal} to exploit the transferability of adversarial examples over models. Although several papers have verified the transferability properties  \cite{madry2017towards}, the focus of our work is on the gradient estimation technique \cite{chen2017zoo} because of the convenience of attack.
This property transferability of adversarial attacks is investigated in \cite{E} for dispersion reduction attack. It uses limited perturbations compared to the existing attacks and demonstrated its performance over different computer vision tasks (image classification, object detection, semantic segmentation). 

The first work to generate adversarial examples for black-box attacks in video recognition, V-BAD~\cite{VBAD} framework utilizes tentative perturbations transferred from image models and partition-based rectifications to obtain good adversarial gradient estimates. They demonstrate an effective and efficient attack with a $\sim$90\% success rate using fewer queries to the target model. 
More recently, the first article on adversarial examples for sign recognition systems in AV \cite{DARTS} has proposed two different attack methods: out-of-distribution and lenticular printing in black-box settings. 


Unlike the scored-based and transfer-based methods, the TRansferable EMbedding based Black-box Attack (TREMBA) method \cite{G}. Direted to an unknown target network, it learns a compact embedding with a pre-trained model and performs an efficient search over the embedding space. The adversarial perturbations by TREMBA have high-level semantics, which is effectively transferable. Further, these perturbations help in enhancing the query efficiency of the black-box adversarial attack across the architectures of different target networks.

The boundary attack is introduced as a category of the decision-based attack \cite{R}, which is relevant for the assessment of model robustness. These are used to highlight the security risks of machine learning systems belonging to closed-source like autonomous cars.
Boundary attacks usually require a large set of model queries for obtaining a successful human indistinguishable adversarial example. To improve the efficiency of the boundary attack, it must be combined with a transfer-based attack. The biased boundary attack \cite{S}, significantly reduces the number of model queries with the combination of low-frequency random noise and the gradient from a substitute model. Similar to other transfer-based attacks, a biased boundary attack depends on the transferability between the target model and the substitute model. The boundary attack++ \cite{T} is an algorithmic improvement of the boundary attack, which estimates the gradient direction with the help of binary information available at the decision boundary. Another method \cite{U} of decision-based attack, called qFool, used very few queries in the computation of adversarial examples. The qFool method can handle both non-targeted and targeted attacks with less number of queries. 

A simple black-box adversarial attack, called SimBA \cite{F} has emphasized that optimizing queries in black-box adversarial attacks continues to be an open problem. This is happening even though there is a significant body of prior work~\cite{VBAD, G}. The algorithm in SimBA repeatedly picks a random direction from a pre-specified set of directions and uses continuous-valued confidence scores to perturb the input image by adding or subtracting the vector from the image. We have extended their work by improving the efficiency and efficacy of the attack. 
Instead of maximizing the loss of the original class, our model searches for gradients in a direction that minimizes the loss of the ``most confused class''.

The main objective of this research is to design black-box adversarial attacks for AV for exposing vulnerabilities in deep learning models. We propose a ``multi-gradient'' attack in deep neural networks model for traffic scene perception. 
There are three main advantages of our model: fast convergence, flattens the confused class probability distribution, and produces adversarial samples with the least confidence in true class. In other words, the results demonstrate that our model is better at generating successful mis-predictions at a faster rate with a higher probability of failure.
Our work in building such models will serve two primary scientific communities. First, it contributes towards the safety and security of the primary users i.e. passengers and pedestrians. Second, it helps AI researchers in developing robust and reliable models.

The main contributions of this work are:
\begin{itemize}
    \item A novel multi-gradient model for designing a black-box adversarial attack on traffic sign images by minimizing the loss of the most confused class.  
    \item Result validation by comparison with transfer-based projected gradient descent (T-PGD) and simple black-box attack (SimBA) using German Traffic Sign Recognition Benchmark (GTSRB) dataset
    \item Our model outperforms on three metrics: iterations for convergence, class probability distribution, and confidence values on input class. 
\end{itemize}

The paper is organized as follows. 
In Section~\ref{proposed_method}, we describe the proposed architecture of black-box adversarial attacks.  Section ~\ref{experiments} contains discussions on the performance of the proposed method on the GTSRB dataset along with quantitative and qualitative analysis. The conclusions are presented and future work in Section~\ref{Conclusion}.

\section{Proposed Method} \label{proposed_method}
\begin{figure*}[htb]
    \centering
    \includegraphics[scale=0.35]{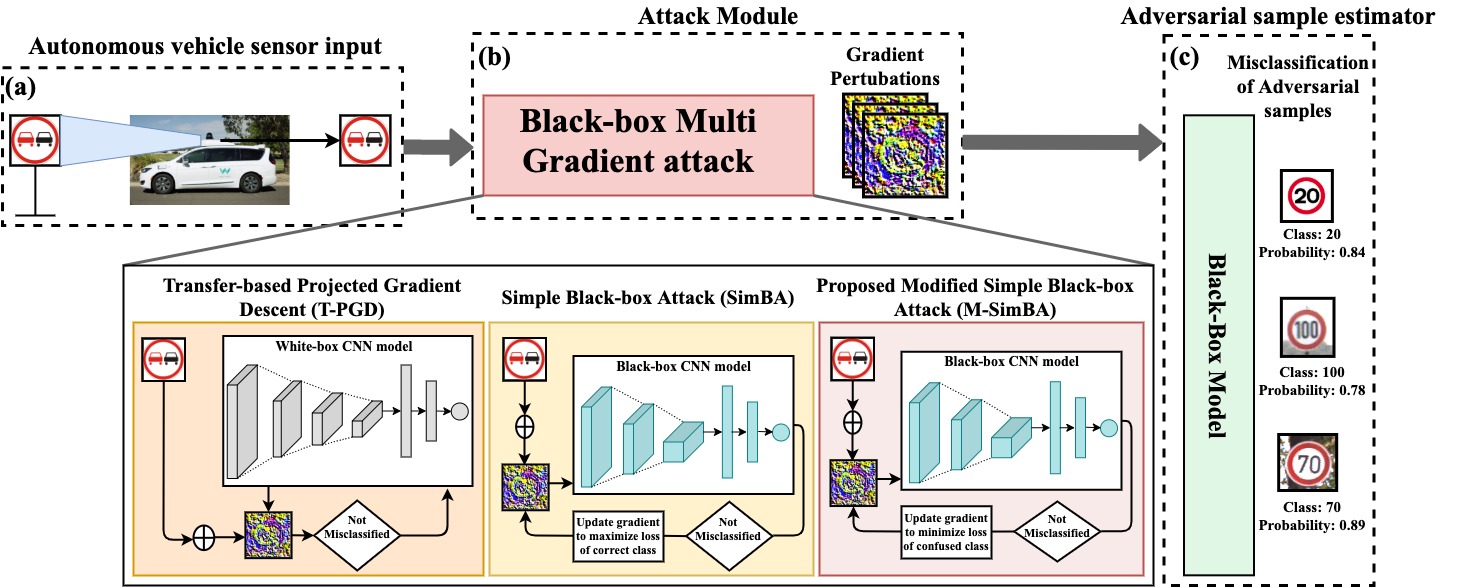}
    \caption{Proposed method for black-box adversarial attacks in autonomous vehicle technology. (a) an input module to sense/detect the traffic signs through the camera attached to the autonomous vehicle (b) multi gradient attack module to generate 3 different gradient perturbations from Transfer based projected gradient descent (T-PGD), Simple Black box attack (SimBA), Modified Simple black-box attack (M-SimBA), and (c) a classification module which attacks the target black-box model }
    \label{fig:proposed}
\end{figure*}

In this section, we are presenting the proposed method for black-box adversarial attacks in AV. As shown in Fig \ref{fig:proposed}, there are three main modules: (a) input module to sense/detect the traffic signs through the camera attached to the autonomous vehicle (b) multi gradient attack module, and (c) adversarial sample estimator that implements the target attack. The gradient perturbations can be generated from one of the three methods: Transfer based projected gradient descent (TPGD), a Simple Black box attack (SimBA), and  Modified Simple black-box attack (M-SimBA). A detailed explanation of this key attack module is given in the subsequent sections.

\begin{figure}[htb]
    \centering
    \includegraphics[scale=0.4]{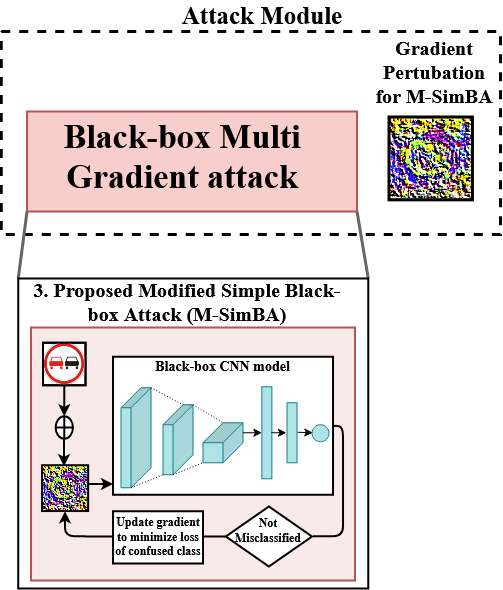}
    \caption{Basic block diagram for Modified Simple Black-box Attack  (M-SimBA)}
    \label{fig:attack_3}
\end{figure}

\subsection{Transfer based Projected Gradient Descent (T-PGD)}
In this white-box attack, the source CNN architecture is trained for a similar task. The gradients from this model are used to produce an adversarial sample which is then transferred to attack the target.
Gradients updates are performed in the direction which \textit{maximizes} the classification loss as per equation (\ref{eq1}), where $x$, $Adv_x$ are original and adversarial sample, respectively.
The term $\epsilon$ is the step size that decides the magnitude of the update. The gradient of the loss function is denoted by $\nabla_x \mathit{J}$ and weights corresponding to the CNN is shown as $\theta$. The output label is shown $y$. 

\begin{equation}
    Adv_{x} = x + \epsilon\ * \ \mathbf{sign}(\nabla_x \mathit{J}(\mathbf{\theta},x,y)).
    \label{eq1}
\end{equation}
Iterative gradient updates are performed until the loss converges to a higher value. This treatment makes the adversarial image to deviate from the original image, making it unperceivable to humans. Although T-PGD shows good generalization ability for samples generated on white box source model to be transferred to the black box model, it is limited by the need for the white box source model.
\begin{figure}[htb]
    \centering
    \includegraphics[width=0.65\linewidth,height=9cm]{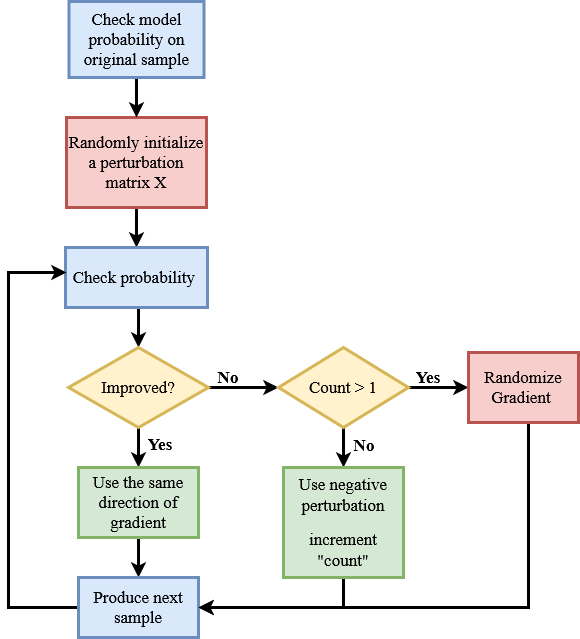}
    \caption{Flowchart of Modified Simple Black-box Attack (M-SimBA)}
    \label{fig:attack_flowchart}
\end{figure}
\subsection{Simple Black-box Attack (SimBA)}
This query-based attack does not require any additional white-box model unlike T-PGD to create the adversarial samples. It has no knowledge of the model and its architecture. Hence, the model parameters such as weights and biases are not known to calculate the gradient concerning the input image as done in previous transfer-based attacks. The SimBA attack uses only the confidence or output probabilities of a black box CNN model to produce adversarial samples.
It tries to search in various directions so that updating the input pixels in that direction \textit{maximizes} the loss of the correct class. This reduces the overall confidence of the network. 

For any given direction $q$ and step size $\epsilon$, 
one of the gradient term $(\mathit{x + q\epsilon})$ or $(\mathit{x - q\epsilon})$ is likely to decrease $P(y | x)$. To minimize the number of queries to the model, $+q\epsilon$ term is added.
In case, this decreases the probability $P(y | x)$, then a step is taken in this direction. Otherwise, the opposite of $-q\epsilon$ is considered. Although it is a simple method to be used to attack any unknown architecture, it requires an extensive gradient search which consumes a large number of iterations to converge.
\begin{figure}[htb]
    \centering
    \includegraphics[scale=0.28]{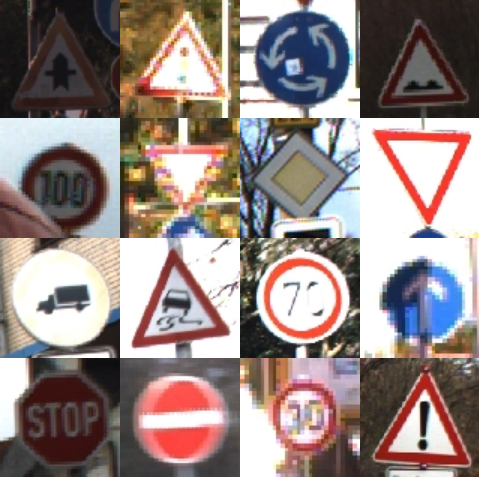}
    \caption{German Traffic Sign Recognition Benchmark (GTSRB) dataset}
    \label{fig:dataset}
\end{figure}
\begin{figure}[htb]
    \centering
    \includegraphics[scale=0.3]{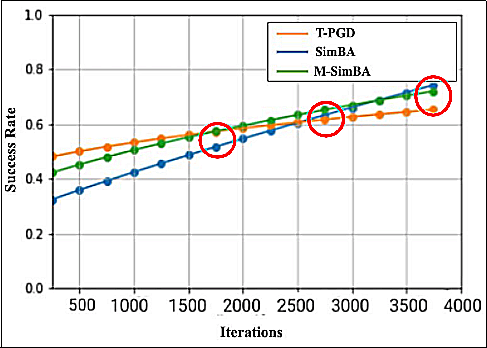}
    \caption{Comparison of three attacks on Iterations vs Success rate}
    \label{fig:comparison_1}
\end{figure}
\begin{figure}[htb]
    \centering
    \includegraphics[scale=0.3]{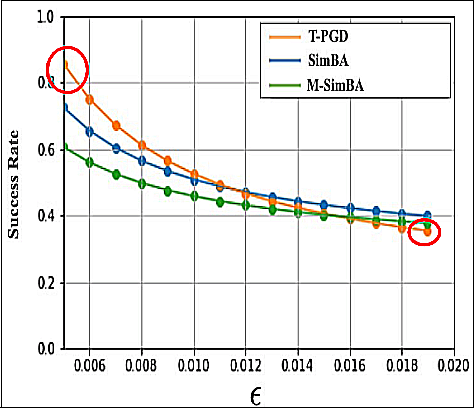}
    \caption{Comparison of three attacks on Epsilon vs Success rate}
    \label{fig:comparison_2}
\end{figure}
\begin{figure}[htb]
    \centering
    \includegraphics[scale=0.3]{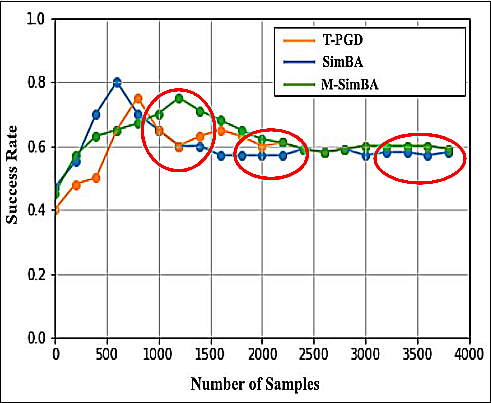}
    \caption{Comparison of three attacks on Samples vs Success rate}
    \label{fig:comparison_3}
\end{figure}
\begin{figure*}[htb]
    \centering
    \includegraphics[scale=0.45]{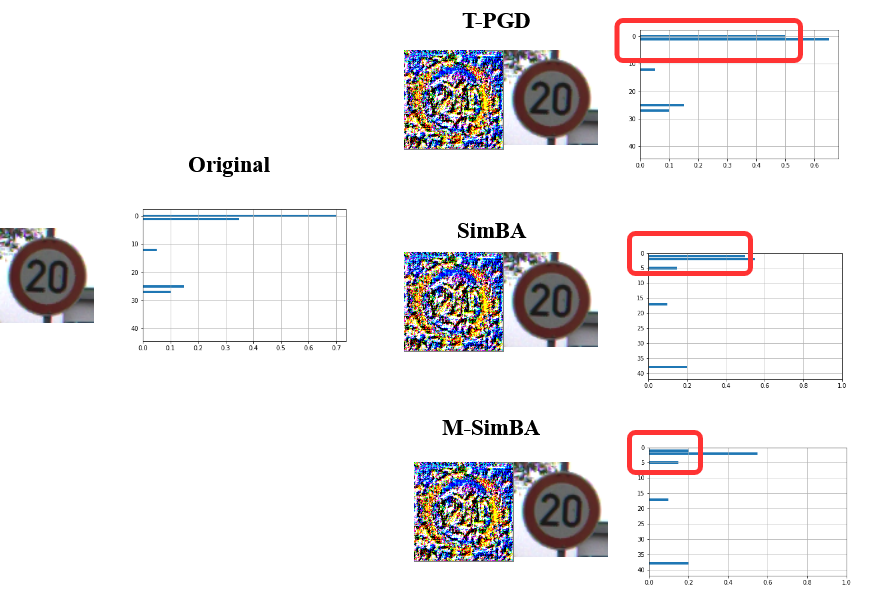}
    \caption{Visual Results on GTSRB - 1. True class of the input image is 0. The T-PGD method produces the adversarial sample highest probability (red box on T-PGD plot) compared to the other two attacks. M-SimBA (red box on M-SimBA plot) can attack the black-box model which outputs very low confidence in the input class i.e., 0. It is a desirable behavior of a robust attack method to suppress the confidence of the original class.}
    \label{fig:visual_1}
\end{figure*}
\subsection{Modified simple black-box attack (M-SimBA)}
To avoid the use of white-box source model of T-PGD attack and late convergence problems of SimBA attack, we are proposing a novel method by modifying the Simple Black box attack to call it M-SimBA. This is shown in Fig. \ref{fig:attack_3}.
Instead of maximizing the loss of the original class in SimBA model, we are \textit{minimizing} the loss of the most confused class. It is the incorrect class where the model misclassifies with the highest probability.
As shown in Fig. \ref{fig:attack_flowchart}, firstly probability of the original model class is checked before the attack. In the next step, random gradients are initialized and are added to the input sample. Subsequently, the black-box model probability is calculated in the most confused class. 
Initially, a positive update is considered. In case, it fails to improve the probability of a most confused class, a negative gradient update is performed. If both positive and negative gradient updates fail to improve the probability, a new gradient is randomly initialized and the process is repeated until convergence.

\section{Experimental results} 
\label{experiments}
In this section, we are presenting the details about the dataset,  experimental setup and result discussions. 

\subsection{Dataset}
We are evaluating the performance of the proposed method on the German Traffic Sign Recognition Benchmark (GTSRB) dataset \cite{GTSRB}. It consists of 43 traffic sign classes, where 39000 are training images and 12000 are test images. The images contain one traffic sign, a border of 10\% around the actual traffic sign (at least 5 pixels) to allow for edge-based approaches. It varies between ($15 \times 15$) to ($250 \times 250$) pixels and sample images are shown in Fig. \ref{fig:dataset}.

\begin{figure*}[htb]
    \centering
    \includegraphics[scale=0.45]{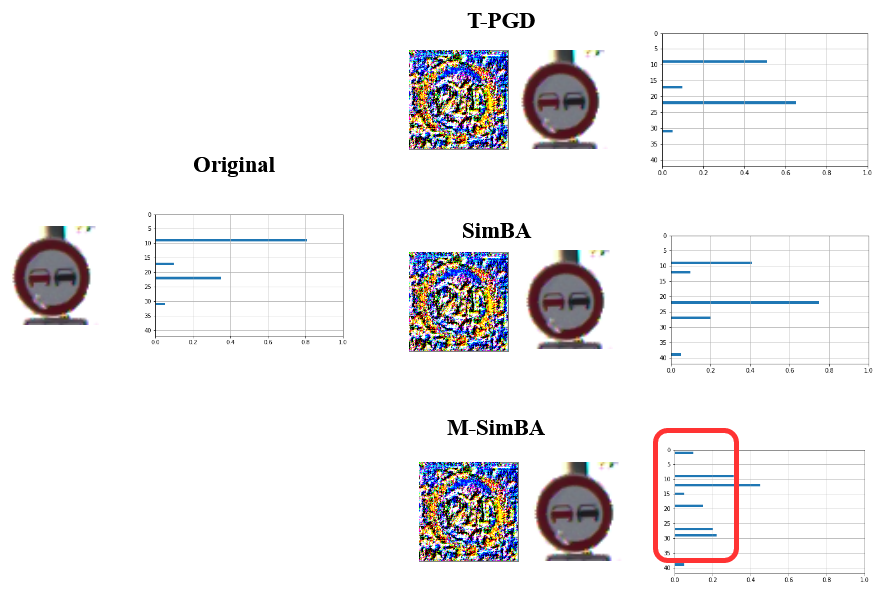}
    \caption{Visual Results on GTSRB - 2. True class of the input image is 9. M-SimBA flattens the distribution of confused class probabilities (red box on M-SimBA plot) compared to the other two attacks. It is a desirable behavior such that there is a high chance that the black-box model confuses with at least of the other class. 
}
    \label{fig:visual_3}
\end{figure*}

\subsection{Experimental Setup}
In this section, we are describing the initial setup for the three models to ensure their proper functioning without attack. 
To perform transfer based projected gradient descent (T-PGD) attack, a 2-layer customized white-box CNN architecture is designed which takes the input image of size (150x150). The model classifies the original samples with 94\% accuracy. It serves as a white-box source to generate adversarial samples in the T-PGD attack. 
To perform SimBA and M-SimBA attack methods, another 2-layer customized black-box CNN architecture with a larger number of max-pool and dropout layers compared to white-box CNN is designed. It takes the input image of same size (150x150) to perform the attack. It classifies the original samples with 96\% accuracy. 

\subsection{Comparison results}
In this section, we are comparing the three attack methods based on their success rate. It is defined as a fraction of generated samples that are successfully misclassified by the black-box model. 
As shown in Fig. \ref{fig:comparison_1}, the success rate increases with an increase in the number of iterations for all the three methods. This is an expected trend, gradient updates for adversarial sample become better with more processing time.
The success rate of T-PGD does not increase much with an increase in iterations, since it does not rely on random searching and requires only a fixed number of iterations to generate the sample.  One of the features of our proposed M-SimBA attack model is that converges faster as compared to the other two methods. 

In the result shown in Fig. \ref{fig:comparison_2}, a common trend is observed that as $\epsilon$ increases, the success rate decreases for all the three methods. This is expected behavior because, as we increase the step size, the value of the gradient update also increases. 
For the large values of $\epsilon$, there is a high probability of overshooting and missing the optimum value. 
Due to this reason, the method may not converge and that can lead to a low success rate. 
On the other hand, T-PGD gives very good results for small values of $\epsilon$, but becomes the poorest of the three methods for larger values of $\epsilon$. This happens as T-PGD relies on gradient updates in a fixed direction and ends up reaching the optimum value in the neighborhood boundary quickly. 
In addition, SimBA and M-SimBA tend to outperform T-PGD and converge to the same point at higher values of $\epsilon$, but SimBA needs a higher number of iterations. 
Finally, in Fig. \ref{fig:comparison_3}, it is observed that M-SimBA tends to show a higher success rate for the initial increase in the number of samples and continues to outperform other methods, because of its property of early convergence. 

\subsection{Qualitative analysis} 
There are two main ideas for the qualitative analysis of the proposed black-box adversarial attacks on GTSRB dataset.
\textit{Firstly}, M-SimBA suppresses the confidence of the original class, which makes it a desirable feature for attack technique.
As shown in Fig. \ref{fig:visual_1}, the true class of the sample is zero. The T-PGD method leads to minimum distortion in the probability vector. On the other hand, M-SimBA can attack the black-box model with very low confidence in the input class with almost zero value. 
%
\textit{Secondly}, M-SimBA flattens the distribution of confused class probabilities compared to the other two attacks as shown in Fig. \ref{fig:visual_3}. This is a advantageous from attack perspective, because it provides a higher chance that the prediction model confuses with the other class.

\section{Conclusion} \label{Conclusion}
Autonomous vehicles powered with deep neural networks for scene perception can be extremely vulnerable to adversarial attacks. For the safety and security of pedestrians and passengers, it is crucial to understand the attacks for building robust models. 
The main objective of our research is to demonstrate and evaluate the black-box adversarial attack for traffic sign detection for AV. 
To achieve efficiency in the iterative process of reducing the number of queries searching the classifier, we focus on minimizing the loss of the most confused class. 
We are comparing our model with two other algorithms SimBA and T-PGD using the GTSRB dataset. 
We are showing the efficiency and efficacy of our model with three different metrics namely:  iterations for convergence, class probability distribution, and confidence values on input class. 
In the future, this work can be extended to attacks in video context and different vehicle sensor data. Also, novel methods can be explored to design robust defense techniques to tackle these adversarial attacks. 

\bibliographystyle{IEEEtran}
\bibliography{workshop}

\end{document}